\documentclass{article}

\usepackage{apalike}
\usepackage{latexsym}
\usepackage{graphicx}
\usepackage{lipsum}
\usepackage{amsmath}
\usepackage{algorithm}
\usepackage{algpseudocode}

\usepackage{amsmath}
\usepackage{amsfonts}
\usepackage{amssymb}
\usepackage{amsbsy}
\usepackage{amsthm}
\usepackage{ragged2e}

\usepackage[pdftex,colorlinks=true,urlcolor=blue,citecolor=black,anchorcolor=black,linkcolor=black,bookmarks=false]{hyperref}

\usepackage{hyphenat}
\newtheoremstyle{scsthe}
{8pt}
{8pt}
{\it}
{}
{\bf}
{.}
{.5em}
{}

\theoremstyle{scsthe}

\begin{document}

\title{Decision Oriented Technique (DOTechnique): Finding Model Validity Through Decision-Maker Context}

\author{
\\
Raheleh Biglari \\
Joachim Denil \\[12pt]
Cosys-Lab, University of Antwerp \\
Flanders Make@UAntwerpen\\
Antwerp, Belgium \\
\{raheleh.biglari,\\
joachim.denil\}@uantwerpen.be\\
}

\date{} 
\maketitle

\section*{ABSTRACT}
Model validity is as critical as the model itself, especially when guiding decision-making processes. Traditional approaches often rely on predefined validity frames, which may not always be available or sufficient. This paper introduces the Decision Oriented Technique (DOTechnique), a novel method for determining model validity based on decision consistency. By evaluating whether surrogate models lead to equivalent decisions compared to high-validity models, DOTechnique enables efficient identification of validity regions, even in the absence of explicit validity boundaries. The approach integrates domain constraints and symbolic reasoning to narrow the search space, enhancing computational efficiency. A highway lane change system serves as a motivating example, demonstrating how DOTechnique can uncover the validity region of a simulation model. The results highlight the technique’s potential to support finding model validity through decision-maker context.

\textbf{Keywords:} Model Validity, Decision-Oriented, Validity Region, Surrogate Models, Decision Consistency, Model-Based Systems Engineering, Simulation Validity, Symbolic Reasoning, Validity Frame, Cyber-Physical Systems.

\section{Introduction}

A critical aspect of using a model effectively is understanding the regions where it remains valid. A model is a representation of a real-world system, designed to operate within specific boundaries and under certain assumptions. These boundaries define the region within which the model can reliably predict outcomes or simulate behaviors. Even when a model performs well within its defined region, applying it beyond those regions introduces significant risks, as its accuracy and reliability are no longer guaranteed. This misapplication can invalidate the model and lead to incorrect results or decisions. For example, a physical model may assume linear behavior within a limited range of forces or temperatures, while a machine learning model may depend on the statistical properties of the training data. When these constraints are violated, for example, by employing the model with extreme input values or in a different context, the model is likely to fail, either by producing incorrect results or becoming entirely unreliable.

Sometimes, this information is already available in the form of validity frames~\cite{van2024validity,van2020exploring}. Validity frames provide predefined constraints, often obtained from domain knowledge, actual facts, or theoretical derivations. 
However, in some cases, the boundaries of a model’s validity are not well-documented or fully understood, leaving users without a clear guide for its appropriate application. This lack of information can lead to significant challenges. Without clearly defined validity frames, users risk using the model in unsuitable contexts, leading to reduced performance and potentially critical errors in decision-making processes.

Addressing the challenge of undefined model validity and an unavailable validity frame is a critical issue in model development. In this section, we present a new technique for overcoming this issue.

\section{Background}

\begin{figure}[htb]
    \centering
    \includegraphics[width=0.63\linewidth]{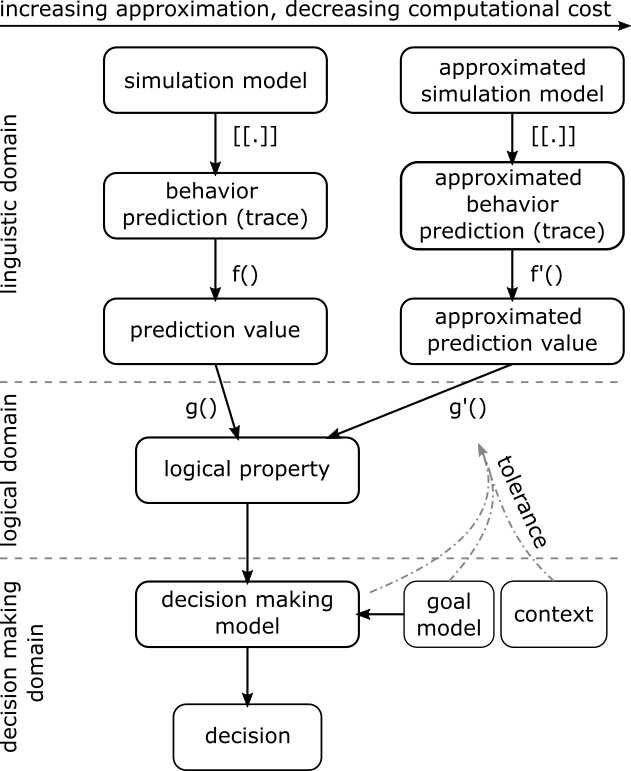}
    \caption{Conceptual framework.}
    \label{fig:concep_frmk}
\end{figure}

We start our conceptualisation inspired by the conceptual framework proposed by Barroca et al.~\cite {barroca2014integrating} depicted in Figure~\ref{fig:concep_frmk}. In our proposed conceptual framework in Figure~\ref{fig:concep_frmk}, we extended the conceptualisation with two different models with different approximations and the layer of the decision-making algorithms. We look into the impact of model approximation and/or abstraction on cost reduction, which forms the foundation of the framework that facilitates system runtime adaptation.
The semantics of a simulation model are determined by simulating the model and using a semantic mapping function ([[.]]), which interprets the model’s behavior by translating its execution traces into meaningful representations, linking the simulation model to its intended real-world interpretation. The simulation results in a collection of traces. While there are a multitude of simulation traces, not all of the provided information is the quantity of interest of the decision-making algorithm or for reasoning over the logical behavior of the system. We need a function $f()$ to extract properties of interest. Then function $g()$ is used to transform between the quantity of interest and the logical property. A logical property is a Boolean value and is on the ontological level, where it gets a real-world meaning. The framework is similar for an approximated/abstracted model and reasons on the same logical property. This means that the function $g'()$ should give an equivalent result to the more detailed model, although with more uncertainty~\cite{biglari2022towards}. As shown in Figure~\ref{fig:concep_frmk}, simulation models produce behaviour traces that influence decision-making through a chain of transformations. When using multiple models of different approximations, our key insight is that model validity can be determined based on decision consistency rather than output similarity.

Traditional validation approaches \cite{oberkampf2010verification} compare model outputs directly, which may be unnecessarily restrictive. Instead, we propose focusing on whether different models lead to equivalent decisions, even if their detailed outputs differ. This aligns with our practical goal: maintaining reliable decision-making while reducing computational costs. 

\section{DOTechnique: Decision Oriented Technique}
In this section, we propose Decision-Oriented Technique (DOTechnique) for establishing model validity when pre-defined validity boundaries are unavailable. 

Let $m_h$ represent the model with the most detail, and $m_s$ the surrogate model under consideration. Let $D$ be the decision maker that maps model outputs to decisions $y \in \mathcal{Y}$, where $\mathcal{Y}$ is the decision space. We define a distance metric $d_Y$ on the decision space:

\begin{equation}
d_Y: \mathcal{Y} \times \mathcal{Y} \rightarrow [0,\infty)
\end{equation}

Since the decision space could be a numerical or categorical (binary/multiple categories/etc.) space, for a given tolerance $\varepsilon > 0$, we define the validity region $\mathcal{V_\varepsilon}$ for numerical decision spaces as:

\begin{equation}
    \mathcal{V_\varepsilon} = \{x \in \mathcal{F} \mid d_Y(D(m_h(x)), D(m_s(x))) < \varepsilon\}
 \end{equation}

 And for categorical decision spaces, we define the validity region as:

\begin{equation}
\mathcal{V_\varepsilon} = \{x \in \mathcal{F} \mid D(m_h(x)) = D(m_s(x))\}
\end{equation}
 
Under the assumption that $\mathcal{V}_\varepsilon$ is continuous, its boundary $\mathcal{B}$ can be characterised as:

\begin{multline}
\mathcal{B} = \{x \in \mathcal{X} \mid \forall \delta > 0, \exists x_1,x_2 \in N(x,\delta) \\
\text{ where } x_1 \in \mathcal{V}_\varepsilon \text{ and } x_2 \notin \mathcal{V}_\varepsilon\}
\end{multline}

where $N(x,\delta)$ represents a ball of radius $\delta$ around $x$. In the rest of the study, we will remove $_\epsilon$ from the notation and implicitly assume that there can be a tolerance associated with the validity region. 

This continuity assumption enables efficient search strategies for identifying the validity boundary. One such algorithm is a simple binary search algorithm that searches for the boundary with a single parameter. However, as most problems are multi-dimensional, this approach does not scale well, and other more feasible algorithms should be employed. Although this topic is related, it is not central to the focus of this paper and, therefore, lies beyond the scope of this research.

\begin{algorithm}
\caption{FindBoundary}
\begin{algorithmic}
\Function{FindBoundary}{$p_1, p_2, \text{tolerance}$}
    \Comment{$p_1 \in \mathcal{V_\varepsilon}, p_2 \notin \mathcal{V_\varepsilon}$}
    \While{$\|p_1 - p_2\| > \text{tolerance}$}
        \State $\text{mid} \gets (p_1 + p_2)/2$
        \If{$\text{mid} \in \mathcal{V_\varepsilon}$}
            \State $p_1 \gets \text{mid}$
        \Else
            \State $p_2 \gets \text{mid}$
        \EndIf
        \EndWhile\\
    \Return $p_1$
\EndFunction
\end{algorithmic}
\label{alg:binarySearch}
\end{algorithm}

\paragraph{Symbolic Reasoning} enables us to leverage domain knowledge and previous experimental results to constrain the search space for valid model regions, e.g., knowledge about physical laws allows us to infer validity regions from individual experimental points. For example, Cederbladh et al. introduce the following example on the braking of a machine: When a test shows that a machine cannot brake safely under certain conditions (e.g., 15220kg at 19 degrees incline), we can deduce that heavier machines or steeper inclines will also fail, without requiring additional experiments. Similarly, successful test cases (like 22330kg at 6 degrees) let us infer validity for lighter machines at lower inclines~\cite{HansreuseExperiment2023}. This systematic approach to knowledge reuse aligns with case-based reasoning principles while incorporating domain-specific constraints to efficiently bound the feasible region. This technique can easily be integrated into our approach. Furthermore, if a partial validity frame with extra constraints is already available we can integrate these constraints as well. 

The search space can be constrained using domain knowledge. Let $\mathcal{C} = \{c_1, c_2, ..., c_n\}$ be a set of domain constraints, where each $c_i: \mathcal{X} \rightarrow \{\text{true}, \text{false}\}$. The feasible region $\mathcal{F}$ is then:

\begin{equation}
\mathcal{F} = \{x \in \mathcal{X} \mid \forall c \in \mathcal{C}, c(x) = \text{true}\}
\end{equation}

where $\mathcal{X}$ is the state space. After defining the feasible region $\mathcal{F}$ based on domain constraints, all validity frame experiments and boundary searching are restricted to points within $\mathcal{F}$. This ensures we only explore physically meaningful states and avoid wasting computational resources on testing invalid configurations. This approach's key advantage is enabling efficient discovery of validity boundaries through search algorithms guided by domain knowledge constraints.

\section{Motivating Example}
To represent the challenges in our research, we use a lane change control algorithm as our motivating example. Highway lane change maneuver system models control the longitudinal and lateral direction of the ego car (target car).

\begin{figure*}
        \centering
        \includegraphics[width =1\columnwidth]{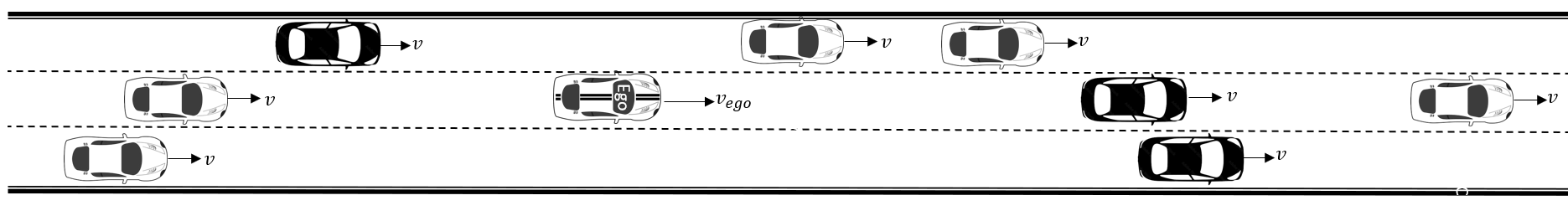}
        \caption{Lane changing scenario.}
        \label{fig:scenarioChpt4}
    \end{figure*}

Figure~\ref{fig:scenarioChpt4} shows a scenario where the front car decelerates slightly, causing the ego car to change lanes. We utilise a Simulink model to simulate this scenario.

To predict the next position of the front car, we look at the front car's next position. In this experiment, we use two different models.
 \begin{itemize}
                 
         \item High-validity model, the most computationally expensive model in this experiment, predicts the trajectories of the current ego vehicle in response to the surrounding vehicles. By incorporating a controller for each vehicle, the model enables a more granular and detailed evaluation of individual car trajectories.

To account for the impact of position changes on trajectory predictions, the model operates under a fixed-point approach, ensuring the emergence of a stable system state. Importantly, the validity of the high-validity model is constrained to the validation domain of the system, aligning its validity frame with the validation domain of the system.
         \item Constant Acceleration model (C.A model) is the kinematic equation $x(t) = 1/2*a*t^2 + v*t + x_0 $. In this formula, $a$ is the vehicle's acceleration in $m/s^2$, $v$, the velocity of the vehicle in $m/s$, and $x$, the vehicle's position in meters.
\end{itemize}

In this experiment, $\mathcal{M}$ is our set of models, where $m_h$ is the high-validity model, $m_{ca}$ is the C.A model. $\mathcal{M}$ =  ($m_h$, $m_{ca}$)

   \begin{figure}[htb]
       \centering
       \includegraphics[width=0.8\linewidth,trim=3cm 9cm 4cm 9cm, clip]{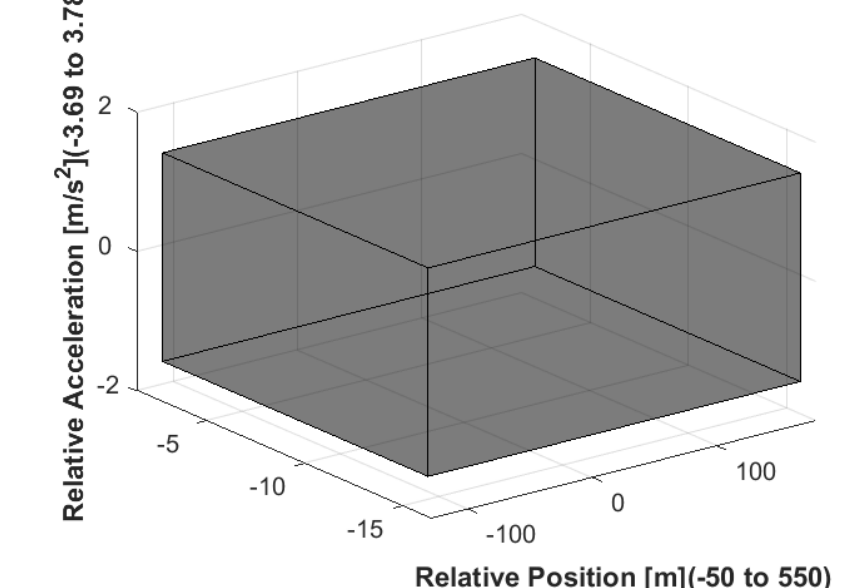}
       \caption{Validity region of the high-validity model.}
       \label{fig:validationDomainTechnique}
   \end{figure}
The validity of $m_h$ is equal to its validation domain shown in Figure~\ref{fig:validationDomainTechnique}. However, the model validity is not pre-defined and available for $m_{ca}$. Therefore,  we use the DOTechnique to find out the model validity.

\begin{figure}[ht]
    \centering
    \includegraphics[width=1\linewidth]{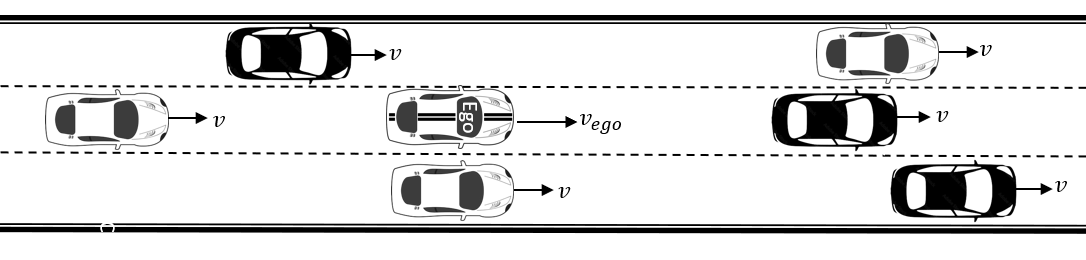}
    \caption{Finding model validity scenario.}
    \label{fig:scenario_VRM_tech}
\end{figure}

To determine the decision-based model validity, we need to consider all possible situations. Therefore, in the highway lane change system with three lanes, we define the scenario depicted in Figure~\ref{fig:scenario_VRM_tech}. It is possible to achieve this by putting six cars surrounding the ego car. The number of lanes, which in this instance is three, is directly proportional to the number of cars that are in the area around them.

To find the validity region $\mathcal{V}$ of $m_{ca}$, we use algorithm~\ref{alg:ValidityRegionSearch}.

\begin{algorithm*}
\caption{ValidityRegionSearch}
\label{alg:ValidityRegionSearch}
\begin{algorithmic}[1]
\Require
    \State $bounds_p = [p_{min}, p_{max}]$ \Comment{position bounds}
    \State $bounds_v = [v_{min}, v_{max}]$ \Comment{velocity bounds}
    \State $bounds_a = [a_{min}, a_{max}]$ \Comment{acceleration bounds}
\Ensure ValidityRegion $VR$
\Procedure{FindValidityBoundaries}{}
    \State $VR \gets \emptyset$
    \ForAll{car $c$ in surrounding\_cars}
        \State $p_{boundary} \gets$ \Call{BinarySearch}{$bounds_p$, matchesDecision}
        \If{$c.position$ is in front}
            \State $valid\_p \gets [p_{boundary}, p_{max}]$ \Comment{Valid when further away}
        \Else
            \State $valid\_p \gets [p_{min}, p_{boundary}]$ \Comment{Valid when further away}
        \EndIf
        
        \ForAll{$p$ in discretise($valid\_p$)}
            \State $v_{boundary} \gets$ \Call{BinarySearch}{$bounds_v$, matchesDecision($p$)}
            \If{$c.velocity$ is positive relative}
                \State $valid\_v \gets [v_{boundary}, v_{max}]$ \Comment{Valid at higher relative speeds}
            \Else
                \State $valid\_v \gets [v_{min}, v_{boundary}]$
            \EndIf
            
            \ForAll{$v$ in discretise($valid\_v$)}
                \State $a_{boundary} \gets$ \Call{BinarySearch}{$bounds_a$, matchesDecision($p,v$)}
                \If{$c.acceleration$ is positive}
                    \State $valid\_a \gets [a_{boundary}, a_{max}]$ \Comment{Valid at higher accelerations}
                \Else
                    \State $valid\_a \gets [a_{min}, a_{boundary}]$
                \EndIf
                \State $VR \gets VR \cup \{(p,v,a) | a \in valid\_a\}$
            \EndFor
        \EndFor
    \EndFor
    \State \Return $VR$
\EndProcedure

\Procedure{BinarySearch}{bounds, decisionCheck}
    \State f$left \gets bounds.min$
    \State $right \gets bounds.max$
    \While{$right - left > \delta$} \Comment{Small delta for numerical stability}
        \State $mid \gets (left + right)/2$
        \If{decisionCheck($mid$)}
            \State $right \gets mid$
        \Else
            \State $left \gets mid$
        \EndIf
    \EndWhile
    \State \Return $mid$
\EndProcedure
\end{algorithmic}
\end{algorithm*}

We check domain constraints $\mathcal{C}$ as the following set:\\
$c_1$: Vehicles maintain deterministic behaviors during the simulation. And in reality, each car might have unknown situations and different behaviour.\\
$c_2$: The minimum car speed has to be 6 $m/s$.\\
$c_3$: Each target vehicle, after acceleration or deceleration, continues with the same velocity.\\
$c_4$: The front gap, the gap between a car and the front car, has to be at least 30 $m$. And the rear Safety Gap must also be 30 $m$.\\
$c_5$: The ego car has no acceleration, and the speed is constant.\\
$c_6$: When the car is in front of the ego car
, the model is valid in positions further away.\\
$c_7$: When the car is behind the ego car 
, the model is valid at positions farther away.

Additionally, for each surrounding vehicle, we determine the boundaries, denoted as 
$\mathcal{B}$, through a binary search process. This involves evaluating various combinations of positions, velocities, and accelerations using the following method:

For each surrounding vehicle, we begin by varying its relative position (with respect to the ego car) using a binary search in Algorithm~\ref {alg:binarySearch}. We then run simulations using both $m_{ca}$ and $m_h$, comparing their outputs to evaluate the predicted trajectories of the ego car. The trajectories do not need to align point-for-point; instead, we define a distance metric $d_Y$ to determine divergence. Specifically, we consider two trajectories as different if they result in a lane change by the ego car. If the trajectories are deemed identical according to $d_Y$, we identify that relative position as the position boundary for the vehicle.

However, identifying the position boundary is not sufficient to establish a valid point within the model. We must also account for whether the vehicle is in front of or behind the ego car. If the vehicle is ahead of the ego car, the feasible region $\mathcal{F}$ extends from the position boundary to the sensor's maximum range (e.g., the end of the highway). Conversely, if the vehicle is behind the ego car, $\mathcal{F}$ is defined as the region between the position boundary and zero.

Next, we determine valid velocities. For each discretised position within $\mathcal{F}$, we explore different velocities using binary search. If the decisions from $m_{ca}$ and $m_h$ align, the point is added to $\mathcal{B}$. This process mirrors the steps used to determine position boundaries. Similarly, we apply the same approach to identify valid accelerations by iterating through all combinations of positions and velocities.

\begin{figure}[htb]
    \centering
    \includegraphics[width=0.8\linewidth]{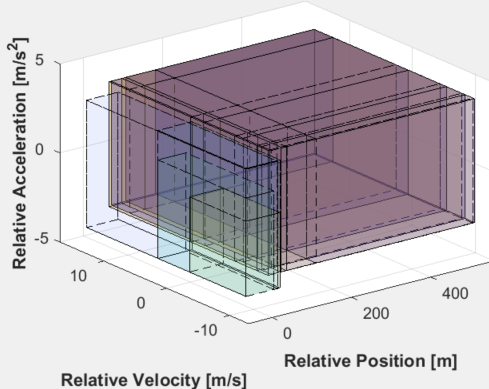}
    \caption{Constant Acceleration (C.A) model validity.}
    \label{fig:vf_CA_tech}
\end{figure}

Therefore, we come up with the following results. The resulting validity region for $m_{ca}$ is illustrated in Figure~\ref{fig:vf_CA_tech}.

\section{Conclusion}
In this paper, we look at the significance of the model's validity, which is equally as crucial as the model itself. To allow this, we propose DOTechnique to find the model validity in the absence of predefined validity in the context of a decision-maker.

By evaluating whether surrogate models yield equivalent decisions to high-validity model, DOTechnique enables the discovery of validity regions even when traditional validity frames are absent.

We acknowledge that a single case study may not provide sufficient validation for broader generalisation, despite the fact that the proposed technique demonstrates promising results in the presented case study. Nevertheless, this highlights the need for future work to further validate and generalise the proposed technique across diverse scenarios and case studies.


\bibliographystyle{apalike}
\bibliography{bibliography}

\begin{thebibliography}{}

\bibitem[Barroca et~al., 2014]{barroca2014integrating}
Barroca, B., K{\"u}hne, T., and Vangheluwe, H. (2014).
\newblock Integrating language and ontology engineering.
\newblock In {\em MPM@ MoDELS}, pages 77--86.

\bibitem[Biglari et~al., 2022]{biglari2022towards}
Biglari, R., Mertens, J., and Denil, J. (2022).
\newblock Towards real-time adaptive approximation.
\newblock In {\em ERTS 2022, June 1-2, 2022, Toulouse, France}, pages 1--5.

\bibitem[Cederbladh et~al., 2023]{HansreuseExperiment2023}
Cederbladh, J., Cleophas, L., Kamburjan, E., Lima, L., and Vangheluwe, H. (2023).
\newblock Symbolic reasoning for early decision-making in model-based systems engineering.
\newblock In {\em 2023 ACM/IEEE International Conference on Model Driven Engineering Languages and Systems Companion (MODELS-C)}, pages 721--725.

\bibitem[Oberkampf and Roy, 2010]{oberkampf2010verification}
Oberkampf, W.~L. and Roy, C.~J. (2010).
\newblock {\em Verification and validation in scientific computing}.
\newblock Cambridge university press.

\bibitem[Van~Acker et~al., 2024]{van2024validity}
Van~Acker, B., De~Meulenaere, P., Vangheluwe, H., and Denil, J. (2024).
\newblock Validity frame--enabled model-based engineering processes.
\newblock {\em Simulation}, 100(2):185--226.

\bibitem[Van~Mierlo et~al., 2020]{van2020exploring}
Van~Mierlo, S., Oakes, B.~J., Van~Acker, B., Eslampanah, R., Denil, J., and Vangheluwe, H. (2020).
\newblock Exploring validity frames in practice.
\newblock In {\em International Conference on Systems Modelling and Management}, pages 131--148. Springer.

\end{thebibliography}

\section*{Author Biographies}

\textbf{Raheleh Biglari} got her PhD in Applied Engineering focusing on Modelling \& Simulation and Digital Twins at the University of Antwerp, Faculty of Applied Engineering in the Department of Electronics and Information and Communication Technologies, Cosys-Lab. Her research interests lie in performance modelling, model-based system engineering, and digital twin engineering.\\
\\
\textbf{Joachim Denil} is an associate professor at the University of Antwerp, Faculty of Applied Engineering in the Department of Electronics and Information and Communication Technologies, Cosys-Lab, and is associated with Flanders Make. His research interests are enabling methods, techniques, and tools to design,
verify, and evolve Cyber-Physical Systems. Specifically, he is interested in performance modelling and
simulation, model-based systems engineering, and verification and validation of models and systems.

\end{document}